\documentclass{article}
\usepackage{spconf,amsmath,graphicx,CJKutf8,tabularx}
\usepackage{url} 

\title{AUTOMATIC RECOGNITION OF SUPRASEGMENTALS IN SPEECH}
\name{Jiahong Yuan$^1$, Neville Ryant$^2$, Xingyu Cai$^1$, Kenneth Church$^1$, Mark Liberman$^2$}
\address{$^1$Baidu Research USA\\
$^2$Linguistic Data Consortium, University of Pennsylvania}
\begin{document}
%
\maketitle
\begin{abstract}
This study reports our efforts to improve automatic recognition of suprasegmentals by fine-tuning wav2vec 2.0 with CTC, a method that has been successful in automatic speech recognition. We demonstrate that the method can improve the state-of-the-art on automatic recognition of syllables, tones, and pitch accents. Utilizing segmental information, by employing tonal finals or tonal syllables as recognition units, can significantly improve Mandarin tone recognition. Language models are helpful when tonal syllables are used as recognition units, but not helpful when tones are recognition units. Finally, Mandarin tone recognition can benefit from English phoneme recognition by combining the two tasks in fine-tuning wav2vec 2.0.
\end{abstract}
\section{Introduction}
\label{sec:intro}

Suprasegmentals are phonological units in speech that are larger than segments (i.e., consonants and vowels), such as syllables, lexical stress, tones, and intonation \cite{leben1973suprasegmental}. In this study, we propose using wav2vec 2.0 \cite{baevski2020wav2vec} fined-tuned with a Connectionist Temporal Classification (CTC) loss \cite{graves2006connectionist} for automatic recognition of suprasegmentals, similar to the approach used by \cite{baevski2020wav2vec} for phoneme recognition on TIMIT \cite{garofolo1993timit}.

Speech segments and suprasegmental units are different not only in the domain of realization, but also in their features. ``Suprasegmental'' has often been used to refer to the phonetic features of suprasegmental units, and used interchangeably with prosodic features. It is well accepted that suprasegmentals are mainly distinguished by pitch, duration, and energy, whereas segments are distinguished by spectral information. There are studies, however, suggesting that spectral information may be also important for suprasegmentals. For example, \cite{Sluijter1996SpectralBA} found that spectral balance is a reliable acoustic correlate of lexical stress, and \cite{Ryant2014MandarinTC,Ryant2014SpeechProsody} demonstrated that MFCCs outperform prosodic features for automatic recognition of Mandarin tones.

With the advent of deep learning and end-to-end models, feature engineering has been largely abandoned with feature representations learned implicitly during training by the neural network. In earlier work, these featured were learned in a supervised fashion from paired audio and transcripts \cite{hinton2012deep}. However, more recent work has focused on unsupervised learning of representations using only audio \cite{oord2018representation,schneider2019wav2vec,baevski2020wav2vec}, which are then used by downstream tasks such as speech-to-text. When compared to conventional acoustic features such as MFCCs, these representations substantially lower the amount of labeled data needed to train state-of-the-art speech-to-text systems for English \cite{baevski2020wav2vec} and low-resource languages \cite{Yi2020ApplyingWT}. Similarly encouraging results have been demonstrate for speaker recognition \cite{oord2018representation} and phone recognition \cite{baevski2020wav2vec}. However, it is not yet clear how well these representations perform for suprasegmental recognition, which requires the network to learn prosodic instead of, or in addition to, spectral features and representations. 

We conducted experiments with fine-tuning wav2vec 2.0 models using CTC for recognition of suprasegmentals, including syllables, tones, and pitch accents. We also made an effort to improve recognition of Mandarin tones by utilizing segmental information. The main results of our study are as follows:

\begin{enumerate}
  \item We demonstrate that fine-tuning wav2vec 2.0 with a CTC loss can improve the state-of-the-art for automatic recognition of suprasegmentals, including syllables, tones, and pitch accents. 
  \item Utilizing segmental information, by employing tonal finals or tonal syllables as recognition units, can significantly improve Mandarin tone recognition. Language models are helpful when tonal syllables are used as recognition units, but not helpful when tones are recognition units.
  \item Mandarin tone recognition benefits from English phoneme recognition by combing the two tasks in fine-tuning wav2vec 2.0.
\end{enumerate}

\section{Related work}
\label{sec:work}

\subsection{wav2vec 2.0}

Wav2vec 2.0 is a framework for self-supervised learning of speech representations. The speech signal is processed by a multilayer convolutional network to obtain latent representations every 25 ms, which are then fed into separate vector quantization and transformer networks. Training is performed using a noise contrastive estimation task in which consecutive sequences of frames of latent representations are masked and the network required to identify the correct quantized representation for each masked time step from a set of distractors sampled from the other masked time steps. This selection is made on the basis of cosine similarity between the quantized representations and the outputs of the transformr network.

Pre-trained wav2vec models can be fine-tuned for speech recognition with labeled data and a CTC loss. \cite{baevski2020wav2vec} demonstrated that this approach achieved 1.8\% word error rate on the test-clean set of Librispeech \cite{panayotov2015librispeech} with a Transformer language model, and 8.3\% phone error rate on TIMIT test set without a language model. \cite{Yi2020ApplyingWT} applied wav2vec 2.0 to speech recognition in low-resource languages. The paper reported more then 20\% relative improvements in six languages compared with previous work. It found that using coarse-grained modeling units,  such as subwords and characters, achieved better results than fine-grained modeling units, such as phones and letters. Fine-tuning wav2vec 2.0 has also been used to perform other tasks such as speech emotion recognition by adding a classification head on the top of the network. \cite{Cai2021Emotion} proposed a multi-task learning framework to simultaneously perform speech recognition and emotion classification with wav2vec 2.0, achieving the state-of-the-art performance on speech emotion recognition.

\subsection{Syllable detection}

Syllables play a crucial role in speech production and perception, and child language acquisition \cite{Cholin2006Syllables, Mehler1981Syllables}. Automatic segmentation of speech into syllables has attracted research interests for decades. \cite{Mermelstein1975AutomaticSO} proposed a method of syllable detection based on “assessment of the significance of a loudness minimum to be a potential syllabic boundary from the difference between the convex hull of the loudness function and the loudness function itself." \cite{Liberman2020Syllables} found that that even simpler methods, based on selecting peaks in a smoothed amplitude contour, also perform quite well on this task. 

Automatic detection of syllables has also been applied for speaking rate estimation. Motivated by the work of counting objects in an image, \cite{Jiao2015Syllables} proposed a more direct way of estimating the speaking rate that does not require segmentation, detection, or peak counting. Their approach achieved a correlation of 0.89 between estimated number of syllables and actual number of syllables on TIMIT test utterances. The SR error rate, i.e., the average of syllable recognition errors across TIMIT test utterances, was 12.2\%. \cite{Sabu2021AnOS} proposed a signal processing pipeline for syllable detection and speaking rate estimation that is optimized based on the direct minimization of naturally arising task-specific objective functions. This approach achieved a correlation of 0.917 and SR error rate of 9.94\% on TIMIT test set.

\subsection{Mandarin tone recognition}

Mandarin Chinese is a tone language. It has four lexical tones, Tone1 to Tone4, plus a neutral tone, Tone5. While the primary acoustic correlate of tones in Mandarin Chinese is fundamental frequency, i.e., F0, other phonetic parameters such as duration, amplitude, vowel quality, etc., are also involved in the production and perception of tones \cite{Whalen1992InformationFM}.

Automatic recognition of Mandarin tones in running speech has been a challenging task due to tonal coarticulation \cite{Xu1997ContextualTV}, tone sandhi \cite{Yuan2014Tone3}, the interaction between tone and intonation \cite{Yuan2004Intonation}, and speaker variation \cite{Moore1997Tone}. \cite{Shih2000ChineseTM} presented an example, for example, showing that the second syllable ying4, which is a lexical falling tone, could be realized as rising in F0 in the phrase of “fan3 ying4 su4 du4” (“reaction time”). The interaction between tones and segments has also been documented in the literature \cite{Tong2008ProcessingDB}.

The employment of deep learning models for Mandarin tone recognition has gained success in recent years. \cite{Ryant2014MandarinTC} built a deep neural network to classify tones in Mandarin Chinese using MFCCs. The system achieved a significant improvement compared to traditional methods using prosodic features on the task, despite the omission of F0 or other pitch-related features. \cite{Lin2018ImprovingMT} studied the effectiveness of articulatory information for Mandarin tone recognition in a DNN-HMM framework. The paper confirmed that the DNN model may be able to extract more useful information from the MFCC parameters for tone recognition. It also found that incorporating the articulatory information into tone modeling can further improve tone recognition, by either explicitly adding the articulatory features or building phone-dependent tonal models. \cite {Lugosch2018ToneRe} propose a method for tone recognition using a convolutional neural network with CTC. This method achieved a tone error rate of 11.7\% on the Aishell-1 dataset. \cite{Peng2021MultiScaleMF} proposed a multi-scale model which can gather information at multiple resolutions to better capture the characteristics of tone variations, achieving competitive results on the Chinese National Hi-Tech Project 863 corpus with TER of 10.5\%. \cite{Tang2021EndtoEndMT} reported that feeding both the Mel-spectrogram and the short term context segment features into an end-to-end model could significantly improve automatic speech recognition, improving the classification accuracy from 79.5\% to 88.7\% on the Aishell-3 database.

\subsection{Pitch accent detection }

In ToBI and ToBI-style intonation transcription, pitch accents and boundary tones are local intonational events and the basic units of intonation \cite{Beckman2005Tobi}. Speakers of English produce certain words in an utterance with special intonational prominence. These pitch-accented words are typically realized with increased duration, intensity, and fundamental frequency. 

The task of automatic pitch accent detection has attracted a considerable amount of research attention. A thorough review of early (prior to 2009) approaches using acoustic and lexical features and a variety of classification models can be found in \cite{Rosenberg2009Tobi}. In two more recent studies, \cite{Stehwien2018Pitchaccents} presented a CNN-based model for this task; and \cite{Nielsen2020TheRO} extended the model to make greater use of context by using full utterances as input and adding an LSTM layer. The studies achieved 87.5\% and 88.7\% accuracy, respectively, on pitch accent detection on American English speech in the Boston University Radio News Corpus.

Most reported studies of pitch accent detection were conducted on the Boston University Radio News Corpus. The corpus contains seven hours of speech from seven speakers, but only subsets of the corpus are labeled with phonetic alignments and intonation markers. It also does not provide a split of train and test sets. Because of these reasons, previous studies based on this corpus have trained and tested on different amounts of data and therefore cannot be easily compared on their performance. For example, Both \cite{Stehwien2018Pitchaccents} and \cite{Nielsen2020TheRO} conducted 10-fold cross validation on the entire dataset, whereas \cite{Ren2004SpeakerIndependentAD} used 78\% of the data from three female speakers only for training and the other 22\% for testing.

\section{Fine-tuning wav2vec 2.0 for automatic recognition of suprasegmentals}
\label{sec:finetuning}

Out procedure for fine-tuning wav2vec 2.0 for suprasegmental recognition is illustrated in Figure 1. The framework is the same as phoneme recognition. In phoneme recognition, phonemes are used as recognition units and the model is fine-tuned using speech waveforms paired with phoneme sequences. In suprasegmental recognition, suprasegmental units are used to replace phonemes as recognition units. A randomly initialized linear projection is added on top of the contextual representations of wav2vec 2.0 to map the representations into suprasegmental units such as syllables, tones, and pitch accents, and the entire model is optimized by minimizing the CTC loss.

\begin{figure}[htb]
  \centering
  \includegraphics[width=1.0\linewidth]{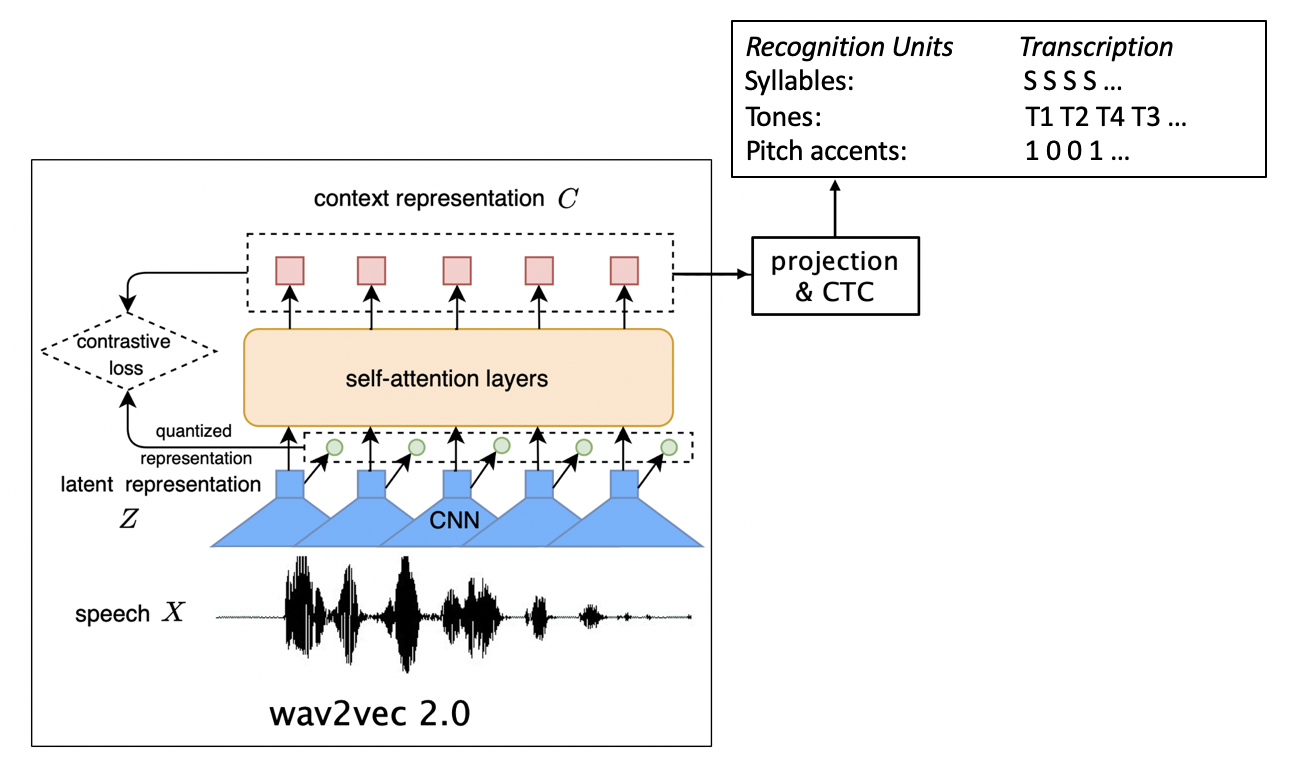}
  \caption{The framework of fine-tuning wav2vec 2.0 for suprasegmental recognition.}
  \vspace{-0.1in}
  \label{fine-tuning}
\end{figure}

Our experiments were conducted using fairseq.\footnote{\url{https://github.com/pytorch/fairseq}} In all experiments, the wav2vec 2.0 large model pre-trained on 960 hours of Librispeech audio (libri960\_big.pt), was used for fine-tuning. For the first 10k updates only the output classifier is trained, after which the Transformer is also updated. The max\_tokens was set to 1.1 million (which is equivalent to 68.75-second audio with sampling rate of 16 kHz), the learning rate was 5e-5. Other details are described below. 

\textit{Syllables:} The TIMIT dataset was used \cite{garofolo1993timit}. The model was trained on TIMIT train set and tested on TIMIT test set. The total number of fine-tuning updates was 20k. The vocabulary contained only one token, `S', which represents a syllable (plus four special tokens that added by fairseq, $<$s$>$, $<$/s$>$, $<$pad$>$, and $<$unk$>$). To generate target labels we simply replaced all vowels in the TIMIT phonetic transcriptions with ``S'' and ignored consonants and other symbols. For example, ``h\# sh ix hv eh dcl jh ih dcl d ah kcl k '' $\rightarrow$ ``S S S S''. Two measures were used to evaluate the performance and compare with previous studies: the correlation between estimated number of syllables and actual number of syllables for utterances in the test set; and the syllable recognition error rate on the test set (because the inference output contains only one type of tokens, `S', there are no substitution errors but only insertion and deletion errors). From the results listed in Table 1, we can see that our approach greatly improved previous results. The correlation was improved from 0.917 to 0.984, and the syllable recognition error rate was reduced by 70\%, from 9.9\% to 2.9\%. 

\textit{Mandarin tones:} Experiments of Mandarin tone recognition were conducted on three datasets: Hub-4 \cite{Hub-4}, Aishell-1 \cite{Bu2017AISHELL1AO}, and Aishell-3 \cite{Shi2020AISHELL3AM}. The Hub-4 dataset is the same as used in \cite{Ryant2014MandarinTC}, which contains 7549 utterances for training and 300 utterances for testing. The utterances were extracted from 20 news announcers in the 1997 Mandarin Broadcast News Speech corpus \cite{Hub-4}. The Aishell-1 and Aishell-3 datasets were downloaded from openSLR. Aishell-1\footnote{\url{https://openslr.org/33/}} contains 165 hours of read speech in Mandarin Chinese from 400 speakers. The speakers are from different dialect regions but most are from northern areas. The corpus includes training (150 hours), development (10 hours), and test (5 hours) sets. Aishell-3\footnote{\url{https://openslr.org/93/}} contains 85 hours of Mandarin speech from 218 native Mandarin Chinese speakers. The corpus has a split of training and test sets, with 174 and 44 speakers, respectively. 

Aishell-3 provides pinyin transcripts. For Hub-4 and Aishell-1, however, only word transcripts are provided. We trained a forced aligner for each of the two datasets, with the Callhome Mandarin Chinese Lexicon \cite{CallhomeLexicon} and the lexicon contained in Aishell-1, respectively, and ran forced alignment to obtain pinyin transcripts for these datasets. The tone marks in the pinyin transcripts were extracted and used as tone labels for the experiments. The vocabulary in the experiments contains five tokens, Tone1 to Tone5. For Aishell-1, its development set was used to find the optimal number of updates. For Hub-4 and Aishell-3, we randomly split the test set into two parts, with one part for development and the other part for testing. We then switched the development and test data to complete testing on the entire test set. 

The results are listed in Table 1. Our approach significantly outperformed previous studies, reducing tone recognition errors by 50\% or more on all three datasets. Table 1 also lists results from two studies on Mandarin Tone recognition using the Chinese National Hi-Tech Project 863 corpus. We don’t have access to this corpus.

\textit{Pitch Accents:} The Boston University Radio News Corpus was used for this experiment. In the corpus, pitch accents are labeled with time stamps but not on words or syllables. Preprocessing is needed to map pitch accent labels to words or syllables using phonetic alignment information. Because not all pitch accents are labeled within a word boundary in the corpus, researchers have used different practices. In \cite{Nielsen2020TheRO} there are 28,489 total word tokens, and 15,544 (54.6\%) of which carry pitch accents, whereas in \cite{Stehwien2018Pitchaccents} there are 26,742 total word tokens, and 13780 (51.5\%) of which carry pitch accents. From our processing of the dataset, we got 30,330 word tokens, and 15,511 (51.1\%) of which carry pitch accents. To compare with previous studies, we conducted 10-fold cross validation on the entire data. In every fold, 10\% of the data was used for testing, 10\% for development, and the remaining 80\% for training. 
In the experiment we first mapped pitch accents to syllables. Every syllable token has a target label of either “1” (pitch accent) or “0” (no pitch accent). Wav2vec 2.0 was, therefore, fine-tuned to recognize two types of syllables, with a pitch accent or without a pitch accent. The inference consists of a sequence of “0”s and “1”s. If any frame within the boundaries of a word has an output of “1” from inference, the word is identified as bearing a pitch accent. The accuracy of pitch accent detection, i.e., the correct identifications divided by the total number of word tokens, is reported in Table 1. We can see that compared to previous studies, our method improved the accuracy from 88.4\% to 89.5\%. 

\begin{table*} [t]
\caption{Results of fine-tuning Wav2vec 2.0 for recognition of syllables, Mandarin tones, and pitch accents, compared to previous studies.}
\label{tab:results}
\begin{center}
\begin{tabular}{|c|c|c|c|c|}
\hline
 & \multicolumn{3}{|c|}{Previous studies} & \\
 Task & Paper & Dataset & Results & Our results\\
\hline
Syllables & Jiao et al. (2015) & TIMIT & Corr: 0.89; SR error rate: 12.2\% & Corr: 0.98; SR error rate: 3.0\% \\
\cline{2-4}
& Sabu, et al. (2021) & TIMIT & Corr: 0.92; SR error rate: 9.94\% & \\
\hline
Mandarin Tones & Ryant et al. (2014) & Hub-4 & Tone error rate: 15.6\% & Tone error rate: 6.0\%\\
\cline{2-5}
& Lugosch, et al. (2018) & Aishell-1 & Tone error rate: 11.7\% & Tone error rate: 5.5\% \\
\cline{2-5}
& Tang\&Li (2021) & Aishell-3 & Classification Acc: 88.7\% & Tone error rate: 6.1\%\\
\cline{2-5}
& Liu, et al. (2018) & HiTec-863 & Tone error rate: 7.2\% & --\\
\cline{2-4}
& Peng, et al. (2021) & HiTec-863 & Tone error rate: 10.5\% & \\
\hline
Pitch accents & Stehwien, et al. (2018) & BURNC & Accuracy: 87.1\% & Accuracy: 89.5\% \\
\cline{2-4}
& Nielsen, et al. (2020) & BURNC & Accuracy: 88.4\% & \\
\hline
\end{tabular}
\end{center}
\end{table*}

\section{Segmental information in Mandarin tone recognition}
\label{sec:segments}

\subsection{Units for tone recognition}

In addition to tones, we also tried two other types of units for recognition of Mandarin tones: initials \& tonal finals (finals+T), and tonal syllables (syllables+T). Examples of transcription using these units are illustrated below:

\begin{CJK*}{UTF8}{gbsn}
\begin{itemize}
  \item Sentence: 她 的 表 现 也 更 加 全 面
  \item Tones: T1 T5 T3 T4 T3 T4 T1 T2 T4
  \item Initials \& finals+T:	t a1 d e5 b iao3 x ian4 ii* ie3 g eng4 j ia1 q van2 m ian4
  \item Syllables+T: ta1 de5 biao3 xian4 ye3 geng4 jia1 quan2 mian4
\end{itemize}
\end{CJK*}
(* “ii” represents a zero-initial as defined in the lexicon of aishell-1) 

For evaluation, the inference output of the model on tones was directly used as the result of tone recognition. The outputs of the other two models were converted to tones for calculating tone error rates. The units of initials in the output were ignored, and tonal finals and tonal syllables were changed to tones by discarding the segmental information, for example, “t a1 d e5” $\rightarrow$ “T1 T5”, and “ta1 de5” $\rightarrow$ “T1 T5”. 

Table 2 lists the results of the three models. The model using tones as recognition units had a tone error rate of 5.5\%. Using initials and tonal finals as recognition units, the error rate was reduced to 5.0\%. The best model used tonal syllables as recognition units, which achieved a tone error rate of 2.8\% on the test set of Aishell-1. 

\begin{table} [t]
\caption{RU and tone error rate of models employing different recognition units.}
\label{tab:tone_units}
\begin{center}
\begin{tabular}{|c|c|c|c|}
\hline
 Recognition   & Vocabulary   & RU           & Tone \\
 Units (RU)    & size         & error rate   & error rate \\
\hline
Tone & 5 & 5.5\% & 5.5\% \\
\hline
Initial \& & 222 & 4.3\% (all RU) & 5.0\% \\
final+T & & 6.3\% (final+T) & \\
\hline 
Syllable+T & 2020 & 4.3\% & 2.8\% \\
\hline
\end{tabular}
\end{center}
\end{table}

\subsection{The effect of language model}

In the experiments above no language models were used in decoding. To study the effect of language models on Mandarin tone recognition, we trained language models using progressively increasing amounts of text data. Three text corpora were used for this purpose:  Aishell-1 word transcriptions, Lancaster Corpus of Mandarin Chinese, and Chinese Gigaword Fifth Edition.

We trained language models of both tones and tonal syllables, to use with wav2vec 2.0 and CTC trained on these units for decoding. To train language models of tonal syllables we converted the Chinese characters in the corpora to pinyin with tone marks, using a Python package called pypinyin\footnote{\url{https://pypi.org/project/pypinyin/}}. The tonal syllables in pinyin were then converted to tones for training language models of tones. For example, “ta1 de5 biao3 xian4 ye3 geng4 jia1 quan2 mian4” is a sample of training data of tonal syllables and “T1 T5 T3 T4 T3 T4 T1 T2 T4” is a sample of training data of tones. 6-gram language models were trained using kenlm, and used for decoding with CTC. Table 3 lists the results in tone error rate for language models trained on different amount of text data, for tonal syllables and tones being recognition units, respectively. 

\begin{table} [t]
\caption{Tone error rate for language models trained on different amount of text data, for recognition units of tonal syllables and tones, respectively. }
\label{tab:language_model}
\begin{center}
\begin{tabular}{|c|c|c|}
\hline
 RU LM (6-gram) & Syllables+T  & Tone\\
\hline
No LM & 2.8\% & 5.54\% \\
\hline
120k sentences & 2.4\% & 5.59\% \\
\hline
1M sentences & 2.1\% & 5.57\% \\
\hline
10M sentences & 1.9\% & 5.56\% \\
\hline
90M sentences & 1.7\% & 5.56\% \\
\hline
\end{tabular}
\end{center}
\end{table}

From Table 3, we can see that language models help tone recognition when tonal syllables are used as recognition units, but they don’t help when the recognition units are tones. With a language model trained on 90M sentences, using tonal syllables achieved a tone error rate of 1.7\% on the test set of Aishell-1, which is dramatically lower than using tones as recognition units (5.5\%). 

\subsection{English phonemes help Mandarin tone recognition}

In this experiment, we explored whether Mandarin tone recognition can be improved by combining it with English phoneme recognition. Our hypothesis is that fine-tuning wav2vec 2.0 to recognize English phonemes will help the network learn representations that are irrelevant to tones because English is a non-tonal language. This may facility the network to better learn representations of tones when fine-tuned to recognize Mandarin tones and English phonemes together.

English TIMIT and Mandarin Hub-4 were used for this experiment. The labels for English TIMIT are English phonemes (we used the canonical phonemes from the TIMIT dictionary, not the transcribed phones), and the labels for Hub-4 are five tones, T1 to T5. The two datasets were combined together for training, with a vocabulary containing both English phonemes and Mandarin Tones. The entire model and vocabulary were used for inference, which is blind to the language identity of test utterances. 

From Table 4 we can see that the tone error rate can be significantly reduced when combined with English phoneme recognition in fine-tuning. Trained on Hub-4 only, the tone error rate was 6.8\% (after 20k updates) on its test set. When combined with TIMIT phoneme recognition, the error rate can be lowered to as low as 4.4\%.

From Table 4 we can also see that the TIMIT phoneme error rate was only 2.3\% when canonical phonemes were used for recognition, which is greatly lower than the error rate of 8.3\% from using transcribed phones (as reported in \cite{baevski2020wav2vec}). This difference is interesting. It may provide insight into why wav2vec 2.0 and CTC perform well on speech recognition, and may suggest a mismatch between speech production and perception, e.g., phonetic reduction vs. perceived deletion. 

\begin{table} [t]
\caption{TIMIT phone error rate and Hub-4 tone error rate in combined fine-tuning.}
\label{tab:language_model}
\begin{center}
\begin{tabular}{|c|c|c|c|}
\hline
 Training data & updates & TIMIT & Hub-4 \\
 & & phone error rate & tone error rate \\
 \hline
TIMIT & 20k & 2.3\% & -- \\
\hline
Hub-4 & 20k & -- & 6.8\% \\
\hline
TIMIT  & 20k & 2.3\% & 5.2\% \\
\cline{2-4}
 \& & 30k & 2.4\% & 5.0\% \\
\cline{2-4}
 Hub-4 & 50k & 2.6\% & 4.4\% \\
\cline{2-4}
 & 80k & 2.7\% & 4.8\% \\
\hline
\end{tabular}
\end{center}
\end{table}

\section{Conclusions}
\label{sec:conclusions}

We demonstrate that fine-tuning wav2vec 2.0 with CTC can improve the state-of-the-art on automatic recognition of suprasegmentals, including syllables, tones, and pitch accents. Compared to previous studies, the method achieved 70\% error reduction on syllable detection, 50\% error reduction on Mandarin tone recognition, and 10\% error reduction on pitch accent identification. 

Segmental information is helpful in Mandarin tone recognition. Employing tonal syllables as recognition units can significantly improve Mandarin tone recognition, compared to using tones as recognition units. Furthermore, language models are helpful when tonal syllables are used as recognition units, but not helpful when tones are recognition units.

Mandarin tone recognition can benefit from English phoneme recognition by combing the two tasks in fine-tuning wav2vec 2.0. The feature space of Mandarin tones is very different from that of English phonemes. Nonetheless, a fine-tuned wav2vec 2.0 model is capable of capturing the inherent characteristics of both tones and phonemes.

\bibliographystyle{IEEE}
\bibliography{mybib}

\end{document}